\documentclass[10pt, a4paper]{article}

\usepackage[final]{lrec2026} 
\usepackage{comment}
\usepackage{multirow}
\usepackage{graphicx}
\usepackage{arydshln}
\usepackage{amsmath}
\usepackage{amssymb}
\usepackage{tcolorbox}
\usepackage{tabularx}
\usepackage{booktabs}
\tcbuselibrary{listings,skins}
\usepackage[normalem]{ulem}

\usepackage{tcolorbox}
\tcbuselibrary{listings,skins}

\newtcolorbox{promptbox}[1]{
  colback=blue!4,
  colframe=blue!50!black,
  boxrule=0.6pt,
  arc=3pt,
  left=6pt,
  right=6pt,
  top=6pt,
  bottom=6pt,
  title=\textbf{#1},
  fonttitle=\small,
  listing engine=listings,
  listing options={
      breaklines=true,
      basicstyle=\ttfamily\small,
  },
}

\useunder{\uline}{\ul}{}
\title{Is Clinical Text Enough?
A Multimodal Study on Mortality Prediction in Heart Failure Patients
}

\name{\large\textbf{Oumaima El Khettari}\textsuperscript{1}, 
      \large\textbf{Virgile Barthet}\textsuperscript{2}, 
      \large\textbf{Guillaume Hocquet}\textsuperscript{3}, \\
      \large\textbf{Joconde Weller}\textsuperscript{3},
    \large\textbf{Emmanuel Morin}\textsuperscript{1},
    \large\textbf{Pierre Zweigenbaum}\textsuperscript{2}}

\address{\textsuperscript{1}Nantes Univ., École Centrale Nantes, CNRS, LS2N, UMR 6004, 44000 Nantes, France \\
\textsuperscript{2}Université Paris-Saclay, CNRS, LISN, Orsay, France \\
        \textsuperscript{3}Hôpital Saint Joseph, DIMID, et Service de Cardiologie, Paris, France \\
        \texttt{\{first.last\}@\{univ-nantes.fr, universite-paris-saclay.fr\}} \\
        \texttt{\{ghocquet,jweller\}@\{ghpsj.fr\}}}

\abstract{
Accurate short-term mortality prediction in heart failure (HF) remains challenging, particularly when relying on structured electronic health record (EHR) data alone. We evaluate transformer-based models on a French HF cohort, comparing text-only, structured-only, multimodal, and LLM-based approaches. Our results show that enriching clinical text with entity-level representations improves prediction over CLS embeddings alone, and that supervised multimodal fusion of text and structured variables achieves the best overall performance. In contrast, large language models perform inconsistently across modalities and decoding strategies, with text-only prompts outperforming structured or multimodal inputs. These findings highlight that entity-aware multimodal transformers offer the most reliable solution for short-term HF outcome prediction, while current LLM prompting remains limited for clinical decision support.
 \\ \newline \Keywords{Clinical NLP, Text classification, Named entity recognition, French medical dataset, Multimodal fusion of text and structured data} }

\begin{document}

\maketitleabstract

\section{Introduction}

Transformer-based models have significantly improved the extraction of semantic representations from clinical text, enabling more effective use of unstructured electronic health records (EHRs) in prediction tasks~\cite{dynomant2019word, liu2023attention}. When fused with structured data, these representations shape a promising multimodal framework, where complementary signals can be combined~\cite{cui2024multimodal,ruan2025evidence}. However, current approaches often focus on global document embeddings and overlook entity-level information when available.

Heart failure (HF) outcome prediction from EHR data has primarily explored unimodal methods, based on structured clinical variables, such as demographics, laboratory tests, and comorbidities; or based on textual notes~\cite{HASHIR2020103489,memarzadeh2022assessingmortalitypredictiondifferent}. Yet in clinical practice, prognostic signals are distributed across heterogeneous modalities: structured variables provide measurable clinical status, while narratives describe patient evolution, symptoms, and treatment context.

Most studies rely on the English MIMIC dataset for patient outcome prediction~\cite{johnson2016mimic}. However, this dataset is not condition-specific, whereas diseases like heart failure may include particular clinical details that are critical for accurately predicting patient outcomes~\cite{nargesi2025automated}. It is also notable that French EHRs differ from the English ones in terminology, note structure and clinical formulation~\cite{d2015redundancy,gaschi-etal-2023-multilingual}. Moreover, conventional text embeddings typically summarize entire documents and overlook entity-level information, which can be decisive for prognosis. These considerations motivate the development of multimodal models that combine structured EHR data with both global and entity-level textual representations, potentially improving the prediction of short-term mortality and rehospitalization risk in HF patients. This work makes the following contributions on a closed French heart failure dataset:
\begin{itemize}
    \item We propose and evaluate a fusion of CLS and entity-level embeddings extracted from clinical notes.
    \item We design and compare several multimodal fusion strategies, combining textual embeddings with structured variables.
    \item We investigate the use of large language models (LLMs) for direct outcome prediction from text, structured data, or both, analyzing performance across modalities.
    \item We perform a systematic comparison of unimodal and multimodal models, evaluating text-only, structured-only, and multimodal approaches.
\end{itemize}
The code for all experiments is publicly available here\footnote{\url{https://github.com/Stan8/heart-failure-multimodal-risk}}.

\section{Related Work}
Numerous models have been proposed to predict heart failure outcomes using structured clinical data. The MAGGIC score uses 13 routinely measured variables and achieves a C-index of 0.74 for 3-year mortality, and index equivalent to ROC-AUC~\footnote{The metric is defined in Section~\ref{sec:expsetup}.} in binary outcome~\cite{rohen2022maggic}. Similarly, the Seattle Heart Failure Model (SHFM) relies on 21 clinical and laboratory variables with ROC-AUC values around 0.73 for 1-, 2-, and 3-year survival~\cite{Li2019-ab}. However, these values reflect only moderate discriminative ability, and the AUROC metric alone provides a limited assessment of clinical utility or performance across different risk levels.

Traditional regression approaches using solely structured variables have also been extensively employed, with Cox proportional hazards models and LASSO regression~\cite{tibshirani1996regression} demonstrating strong predictive capabilities~\cite{assegie2022prediction}. Recent studies show that LASSO Cox regression achieves the highest C-indices of 0.83 for 6-year mortality prediction in heart failure with mildly reduced ejection fraction~\cite{zhao2022machine}. However, performance tends to decline markedly when predicting shorter term outcomes, such as 3 month mortality, like in our case. 

Transformer-based language models have advanced the representation of clinical texts for predictive modeling~\cite{kim2025leveraging,MAO2023104442,rasmy2021med}. Models such as ClinicalBERT, pretrained on MIMIC-III clinical notes, have demonstrated superior performance compared to general-domain BERT in hospital readmission prediction (ROC-AUC = 0.714)~\cite{huang2019clinicalbert}.

Multimodal frameworks integrating structured data with unstructured text have shown promise for clinical prediction tasks such as mortality, readmission, and diagnosis, using different fusion strategies. \citet{zhang2020combining} extract features from structured and unstructured data independently, and use logistic regression, random forest, CNN and LSTM as models. The best results are obtained when fusing unstructured clinical notes, temporal signals, and static information. \citet{cui2024multimodal} introduce the MINGLE framework, an approach that integrates structured clinical records and unstructured clinical notes using large language models and a hypergraph neural network, combining both medical concept semantics and clinical note semantics for improved patient outcome prediction. 

Different fusion strategies exhibit varying performance characteristics. Early fusion (input-level) remains the most commonly used approach~\cite{guarrasi2025systematic}, while intermediate fusion methods using channel-based feature blocks capture global contextual information and inter-modal correlations. Attention-based fusion mechanisms demonstrate superior performance, with multimodal transformers achieving ROC-AUC of 0.877 for in-hospital mortality prediction~\cite{lyu2023multimodal,LIU2023104466}. As underlined in~\cite{guarrasi2025systematic}, few works provide systematic comparisons across multiple fusion strategies, with most studies focusing on single fusion approaches. 

Recently, instruction-tuned large language models have emerged as powerful means for clinical reasoning and decision support. GPT-4 demonstrates the ability to mimic clinical reasoning processes without sacrificing diagnostic accuracy~\cite{savage2024diagnostic}. On the task of disease prediction and readmission, \citet{ben2024cpllm} have shown that LLMs surpass SOTA models without requiring pretraining on medical data using Llama2 13B and BioMedLM 2.7B. However, the direct use of LLMs for patient outcome prediction, particularly when conditioned on different modalities (structured data, clinical text, or both), remains largely unexplored.
\section{Data and Resources}
This section presents the multimodal dataset derived from a French hospital, describing the cohort characteristics, the structured and unstructured data used, and the annotation procedures, combining automated extraction and manual validation of medical entities.

\subsection{Data Collection and Cohort Description} 

The dataset in this study was collected from the cardiology department of Hospital Paris Saint-Joseph in Paris, France. It includes the records of patients hospitalized for heart failure. Each record corresponds to a single stay at the hospital and contains free-text clinical notes written by physicians in French. 

\begin{figure}[!hbt]
\begin{center}
\includegraphics[width=\columnwidth]{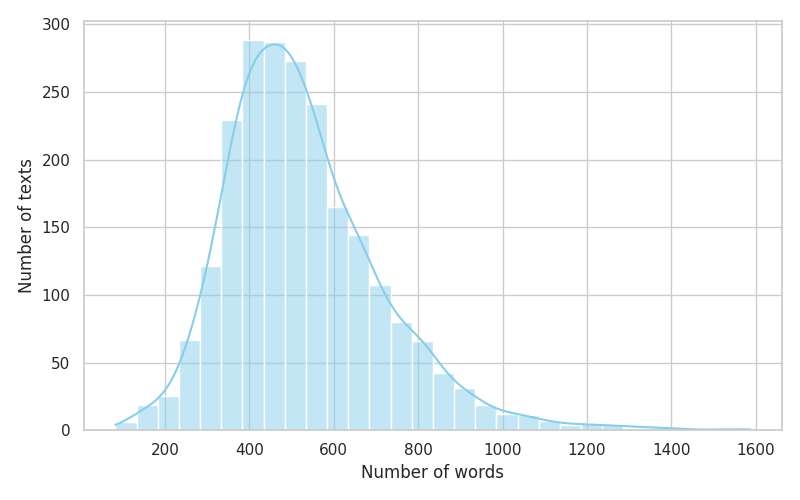} 
\caption{Distribution of clinical notes per size.}
\label{fig:size-distribution}
\end{center}
\end{figure}

Each record is associated with a binary label indicating whether the patient died within three months (called positive cases) following hospital discharge. The final cohort comprises 2,254 patient stays, with approximately 11\% of the patients belonging to the positive class (died within three months) and 89\% to the negative class, illustrating a strong class imbalance typical of real-world clinical datasets. The distribution of word count in the clinical records is shown in Figure~\ref{fig:size-distribution}.

All data were fully de-identified prior to their use, and are accessed exclusively via a secure connection through the hospital's infrastructure. No identifiable personal information was used at any point.
\subsection{Structured Data}

The structured dataset includes 115 demographic, clinical, biological, and therapeutic variables collected at hospital admission. The median age of patients was 81 years, reflecting an elderly population typical of heart failure. Variables cover comorbidities (hypertension, diabetes, atrial fibrillation), laboratory values (BNP, creatinine), cardiac function indicators (LVEF and etiology), and medication use (diuretics, beta-blockers, RAS inhibitors, etc.). After preprocessing and feature selection, this set was reduced to 41 variables used for model development. The complete list of the reduced features and their corresponding categories is provided in Appendix~\ref{app:features}.

\subsection{Clinical Notes}

Clinical records contain free-text notes written during the patient's hospital stay. These notes were taken through the \textit{DxCare} electronic health record system, which structures the text fields in titled sections (patient's background, current treatments, physiological measurements at admission...).

\subsection{Automatic Pre-annotation with Named Entity Recognition and Relation Extraction}

\subsubsection{Automatic heuristic-based preannotation}
An initial pre-annotation was performed using a heuristic-based system combining pattern matching, domain-specific lexical resources as well as syntactic analysis~\cite{barthet-etal-2023-taln}.
The system was built to automatically identify candidate spans corresponding to 23 predefined clinical entity types.
Heuristics relied on: 
\begin{itemize}
    \item \textbf{Exact-term matching} using medical terminologies like the UMLS \cite{BodenreiderNAR2004} or the french public drug database BDPM\footnote{\url{https://base-donnees-publique.medicaments.gouv.fr/}} . 
    \item \textbf{Regular expressions} to detect and label numerical expressions. (\textit{42cm²} $\rightarrow$ Value, \textit{11-04-2016} $\rightarrow$ Date).
    \item \textbf{Morphoyntactic patterns} through prefixes and suffixes (\textit{-pathie} $\rightarrow$ Pathology, \textit{-tomie} $\rightarrow$ Treatment[Interventional]).
    \item \textbf{Trigger word lists} (\textit{Syndrome de} $\rightarrow$ Pathology, \textit{Greffe de} $\rightarrow$ Treatment[Interventional]).
\end{itemize}
To assess the quality of the heuristic-based pre-annotation, a detailed evaluation was conducted using the \textit{brat-eval} tool, comparing automatic annotations to three documents fully annotated by a medical expert. The evaluation was carried out in two conditions:  \begin{enumerate}
    \item \textit{EXACT} type matching, requiring identical entity labels between the gold standard annotation and system annotation.
    \item \textit{OVERLAP} boundary matching, allowing partial span overlap. For example, if \textit{L'insuffisance cardiaque} is labeled as a Pathology in the gold standard annotation, and only \textit{insuffisance cardiaque} is labeled as a Pathology in the system annotation, the prediction is still considered correct.
\end{enumerate}
The overall results indicate a strong recall for key medical entities such as Pathology and Treatment, with a precision above 0.85 for most entity types. 
Lower precision was observed for abstract or contextual entities such as Evolution or Change of location. 

\subsubsection{Manual annotation}

\paragraph{Expert-driven manual annotation}
A subset of 10 texts from the pre-annotated corpus was manually corrected by a cardiology expert from the hospital. The annotator revised the pre-annotated files by adding missing entities, correcting mislabeled ones, and removing irrelevant false positives. This process ensured the clinical validity of the final annotations of those 10 texts.
Beyond producing a high-quality subset, the expert review also helped us improve the quality of the heuristic-based pre-annotation tool through an iterative back-and-forth process: the expert feedback was used to refine the heuristic rules and improve the overall performance of the system.
\paragraph{Non-expert-driven manual annotation.}
In addition to the expert review, a larger subset of 50 clinical notes was manually corrected by a non-expert annotator. 
To assess annotation consistency and reliability, the non-expert’s work was compared against the expert-corrected gold standard. 
For this purpose, the non-expert independently annotated the same 10 documents reviewed by the expert, and the two resulting versions were compared to compute inter-annotator agreement, results to be provided in appendix.
This procedure provided an empirical estimate of the annotation reproducibility between domain experts and non-experts.
\subsubsection{Automatic NER-based annotation}
Following the above work, a supervised Named Entity Recognition (NER) system was developed to extend the annotation coverage and improve consistency across the corpus. A BERT model was fine-tuned on a mixed dataset combining heuristic pre-annotations and manually corrected annotations produced by the non-expert annotator. This hybrid quality of data was designed to balance annotation quality and volume of data, as manually corrected annotations were limited in number but considered of a better quality, while heuristic-based pre-annotations provided a larger corpus but also noisier data.

Several training configurations were tested using different proportions of heuristic and manually corrected data~\cite{barthet:hal-04780743}. 
A configuration combining a subset of higher confidence heuristic annotations with the non-expert corrected subset was found to yield the best F1.

To ensure a fair evaluation of the NER-based annotation system, the 10 texts manually corrected by the expert annotator were excluded from the NER model's training set. This allowed us to 
evaluate the supervised NER model against
the expert-produced gold standard annotation.
The results 
are shown in Table \ref{tab:eval_ner}.
Overall, precision and recall are good or satisfactory in most categories.  The lowest two (\emph{Evolution} and \emph{Duration}) are fairly rare.

\begin{table}[h!]
\centering
\begin{tabular}{lccc}
\hline
\textbf{Label} &\textbf{Precision} & \textbf{Recall} & \textbf{F1} \\
\hline
Age & 1.00 & 1.00 & 1.00 \\
Social circle & 0.87 & 1.00 & 0.93 \\
Time & 1.00 & 0.80 & 0.89 \\
Mode & 0.86 & 0.86 & 0.86 \\
Treatment & 0.90 & 0.82 & 0.86 \\
Dose & 0.77 & 0.92 & 0.84 \\
Negation & 1.00 & 0.73 & 0.84 \\
Autonomy & 0.68 & 1.00 & 0.81 \\
Behavior & 0.67 & 1.00 & 0.80 \\
Concentration & 0.78 & 0.82 & 0.80 \\
Frequency & 0.75 & 0.83 & 0.79 \\
Date & 0.87 & 0.70 & 0.78 \\
Measur. param. & 0.77 & 0.79 & 0.78 \\
Disease & 0.79 & 0.76 & 0.77 \\
Sign/Symptom & 0.80 & 0.74 & 0.77 \\
Location & 0.70 & 0.81 & 0.75 \\
Hypothesis & 1.00 & 0.58 & 0.73 \\
Location change & 0.67 & 0.80 & 0.73 \\
Anatomy & 0.87 & 0.58 & 0.70 \\
Value & 0.67 & 0.69 & 0.68 \\
Examination & 0.59 & 0.63 & 0.61 \\
Evolution & 0.50 & 0.50 & 0.50 \\
Duration & 0.67 & 0.33 & 0.44 \\
\hline
Micro-average & 0.81 & 0.76 & 0.78 \\
\hline
\end{tabular}
\caption{Evaluation of the supervised NER results against expert annotations}
\label{tab:eval_ner}
\end{table}


\section{Methods}

We address short-term mortality prediction in patients hospitalized for heart failure. 
Given the information available at discharge, the objective is to predict whether a patient will die within three months following hospital discharge.

We formulate this problem as a binary classification task. Each admission is labeled as:
\begin{itemize}
    \item $y = 1$ if the patient died within three months after discharge,
    \item $y = 0$ if the patient was alive at three months.
\end{itemize}

Three main families of models are employed to evaluate the contribution of different information sources to mortality prediction in heart failure: (i) text-only models, (ii) structured-only models, and (iii) multimodal fusion models combining both. In addition, we explore a generative formulation of the task, where an LLM is prompted to predict the patient outcome based on the clinical notes, the structured data, or both modalities.

\subsection{Text-Only Methods}
In the text-only methods, we investigate whether leveraging named entities extracted from clinical notes improves mortality prediction compared to using the CLS embedding alone, which is the final hidden-state representation of the prepended [CLS] token in a Transformer model, which aggregates contextual information from all tokens through self-attention and is commonly used for sequence-level classification.
The CLS-only baseline serves as a reference, against which we compare alternative strategies that combine the CLS embedding with entity-level representations.
\paragraph{CLS-only baseline:} For each clinical note, the contextual representation of the [CLS] token is extracted from pretrained French biomedical language models, CamemBERT-bio~\citelanguageresource{touchent2023camembert} and DrBERT~\citelanguageresource{labrak2023drbert}. A linear classifier was trained on top of these embeddings using cross-entropy loss with class weighting.
\paragraph{Entity-based representations:} To better exploit annotation information in clinical text, we incorporated embeddings of annotated medical entities, covering categories such as Anatomy, Treatment, Examination, and Social circle. Each entity embedding was obtained by averaging the subword embeddings within the entity span. When a clinical note contained multiple entities of the same type, a single embedding for that type was computed by averaging the embeddings of all entities belonging to it. As a result, each clinical note is represented by a CLS embedding along with 23 entity-type embeddings. Various fusion strategies were applied to combine the CLS embedding with the entity-level embeddings, to assess the impact of named entities compared to the CLS alone:
\begin{description}
    \item[Average fusion:] For each note, the embeddings of all entities are averaged across entity types, producing a single entity-level vector. This vector is then averaged with the CLS embedding to form the combined representation. 
    \item[Sum fusion:] Similar to average fusion, but the entity-level vectors are summed before being added to the CLS embedding.
    \item[Concatenation:] The mean entity embedding is concatenated with the CLS embedding along the feature dimension.
\end{description}  

These first three fusion strategies are basic strategies to integrate entity-level information with the CLS embedding. Average fusion emphasizes the overall contribution of entities while keeping the same dimensionality as the CLS-only representation. Sum fusion allows the magnitude of multiple entities to proportionally influence the combined representation. Concatenation preserves the CLS information while explicitly adding entity-level information, resulting in an increased input dimensionality. The classifier is, in all these cases, a logistic regression.

Beyond these basic fusion operations, two more strategies were considered to allow the model to learn the relative importance of entity-level information, using a simple linear classification layer. 
\begin{description}
    \item[Weighted fusion] introduces learnable scalar weights $w_{CLS}$ and $w_E$ to balance the contribution of CLS and entity embeddings:
    \vspace{-1em}
    \[
        \mathbf{z} = w_{CLS} \mathbf{h}_{CLS} + w_E \mathbf{h}_E, \quad \mathbf{h}_E = \frac{1}{E}\sum_{i=1}^{E} \mathbf{h}_i
    \]
    \vspace{-2em}
    \item[Gated fusion] uses a trainable gating function $g(\cdot)$ to dynamically modulate the influence of entity embeddings per patient:
    \[
        \mathbf{z} = \mathbf{h}_{CLS} + g(\mathbf{h}_{CLS}, \mathbf{h}_E) \odot \mathbf{h}_E
    \]
    where $g(\cdot) \in [0,1]^d$ is a sigmoid-based gate and $\odot$ denotes element-wise multiplication.
\end{description}

\subsection{Structured-Only Models}
To assess the predictive value of structured data independently from clinical text, two approaches were investigated:
\paragraph{Reference model:}Following the approach adopted in \cite{weller2025prediction}, a comprehensive pipeline was implemented using the full set of 115 structured variables. After preprocessing (missing-value imputation and z-score normalization), a Least Absolute Shrinkage and Selection Operator~\cite{tibshirani1996regression} (LASSO) regression was applied to perform feature selection, retaining the 41 most predictive variables for mortality. The selected subset was then used to train a logistic regression classifier with L2 regularization and class weighting to mitigate imbalance. Hyperparameters were optimized via grid search within a 5-fold stratified cross-validation scheme.
\paragraph{Baseline model:}To provide a more compact and interpretable benchmark, a logistic regression model was trained on the selected subset of 41 structured features. Missing values were imputed with the mean and all variables standardized.

\subsection{Multimodal Fusion Models}
Several multimodal strategies were evaluated to combine textual and structured information, where the textual input could consist of the CLS embedding alone or the combination of CLS and entity embeddings:
\paragraph{Simple fusion methods:} These approaches combine textual and structured representations without explicit interaction or learnable fusion mechanisms:
\begin{itemize}
    \item \textbf{Baseline:} Direct concatenation of textual and all structured representations (CLS+All) or only the selected one (CLS+Sel.), followed by a logistic regression classifier.
    \item \textbf{Late fusion:} Combines the prediction probabilities of text-only and structured-only models through simple averaging or stacking using a linear classifier.
\end{itemize}

\paragraph{Learnable interaction-based methods:} These approaches introduce trainable components to adapt the contribution of each modality :
\begin{itemize}
    \item \textbf{Gated Fusion:} Learns a gating mechanism to modulate the contribution of entity-level embeddings before concatenation with structured data, followed by a multilayer perceptron (MLP) classifier.
    \item \textbf{Dual Cross Attention (Dual-XAttn):} Employs bidirectional cross-attention blocks enabling the CLS and structured representations to attend to each other before classification.
    \item \textbf{Gated Dual Cross Attention (Gated Dual-XAttn):} Combines entity-level gating with cross-attention between textual and structured modalities to dynamically balance textual features and structured variables.
    \item \textbf{Gated Dual Bi-directional Cross Attention (Bi-Gated Dual-XAttn):} Extends the previous model by allowing fully symmetric attention flow between text (CLS + entities) and structured data.
\end{itemize}

\begin{table*}[!ht]
\centering
\begin{tabular}{l|cccc|cccc}
\hline
\multirow{2}{*}{\textbf{Methods}}  & \multicolumn{4}{c}{\textbf{CamemBERT-bio}}                       & \multicolumn{4}{c}{\textbf{DrBERT}}                              \\
                   & \textbf{P}       & \textbf{R}       & \textbf{F1}               & \textbf{AUC}              & \textbf{P}      & \textbf{R}       & \textbf{F1}               & \textbf{AUC}              \\
\hline
Baseline: CLS only & 28.92  & 48.77  & 36.24           & 75.09           & 28.70  & 29.02  & 28.79           & 67.85           \\
Average fusion     & 36.88  & 40.33  & 38.42           & 74.38           & 31.94  & 34.68  & \textbf{33.20}           & \textbf{71.97 } \\
Sum fusion         & 36.49  & 33.92  & 34.79           & 73.97           & 37.02  & 29.91  & \uline{32.60}           & 70.68           \\
Concatenation      & 31.19  & 48.78  & 37.99           & 74.90           & 33.15  & 27.03  & 29.74           & \uline{71.40}           \\
\hdashline
Weighted average   & 32.77  & 47.56  & \uline{38.74}           & \textbf{75.82}           & 31.67  & 26.22  & 28.50           & 70.86           \\
Gated fusion       & 37.07  & 42.33  & \textbf{39.50}           & \uline{75.37}           & 30.12  & 25.39  & 27.24           & 67.66           \\
\hline
\end{tabular}%
\caption{Performance of text-only methods. The best F1 and ROC-AUC results are bolded, and the second best are underlined}
\label{tab:text-results}
\end{table*}

\subsection{Generative LLM-Based Experiments}
To evaluate the ability of LLMs to accurately predict patient outcomes based on different input modalities, we prompted three LLMs, Mistral-7B-Instruct-v0.3~\citelanguageresource{jiang2024mixtral}, Qwen2.5-7B-Instruct~\citelanguageresource{qwen2025qwen25technicalreport} and MedGemma 4B instruct~\citelanguageresource{sellergren2025medgemma} to generate the patient outcome (survival or deceased) based on different input modalities. Prompts contained either (i) the clinical note text, (ii) structured data features, or (iii) a combination of text and structured data, along with an instruction in French asking for the class of the patient based on the input, that is the following: 

\begin{promptbox}{Prompt Used for LLM Mortality Prediction}
You are an expert cardiologist. Based solely on the clinical file (or structured data) provided below, determine whether the patient will die within three months after discharge (class 1) or survive (class 0). Output only the class label (0 or 1).
\end{promptbox}

In addition, to assess whether the predictive performance depends on the format of structured data rather than the underlying values, we transformed the tabular variables into textual summaries using a predefined template. These textified representations were then provided to the LLMs as input for patient outcome prediction, using the same instruction as the previous experiments.
\section{Experimental Setup}
\label{sec:expsetup}

Experiments were evaluated using a stratified 5-fold cross-validation procedure, ensuring consistent mortality prevalence across folds. Reported performance metrics correspond to the mean across the five folds.

For models using pretrained text embeddings, gradients were not propagated through the language model; only the classifier parameters were updated.
Logistic regression models were trained with class weighting to mitigate label imbalance. 

For text generation, we adopted greedy decoding for all LLMs to ensure reproducible and deterministic predictions by selecting the most probable token at each generation step~\cite{shi2024thorough}. Additionally, for Qwen, we evaluated constrained decoding, where the output vocabulary is restricted to the tokens ``0'' and ``1'', forcing the model to produce a valid class label~\cite{tu-etal-2024-unlocking}.

Model performance is evaluated using precision, recall, and F1-score computed on the positive (deceased) class. In addition, we report the Area Under the ROC Curve (ROC-AUC):

\[
\begin{aligned}
\mathrm{ROC\text{-}AUC} &= \int_0^1 \mathrm{TPR}(\mathrm{FPR}) \, d(\mathrm{FPR})\\
             &= \mathbb{P}\big( s(x^+) > s(x^-) \big),
\end{aligned}
\]
where \( s(\cdot) \) denotes the predicted score, and the probability is taken over pairs of positive \( (x^+) \) and negative \( (x^-) \) samples. 
ROC-AUC was not used for LLM evaluation, since their discrete text outputs do not allow threshold variation as in standard probabilistic classifiers.

\paragraph{Statistical Significance:} To assess whether observed differences between the best models are robust, we conducted paired statistical testing across the five cross-validation folds. For each model comparison, fold-level AUC values computed from out-of-fold predictions were compared using a paired t-test. We additionally report 95\% confidence intervals for the mean difference obtained via bootstrap resampling over folds.
\section{Results and Discussion}
\subsection{Text-only methods}

Table~\ref{tab:text-results} presents the performance of text-based approaches using CamemBERT-bio and DrBERT embeddings. Overall, text representations result in moderate discriminative performance, with ROC-AUC values in [70\%, 76\%]. This confirms that clinical notes alone contain impactful prognostic patient-level information for 90-day mortality prediction.

Across all fusion strategies, CamemBERT-bio consistently outperformed DrBERT, achieving higher F1 and AUC scores. This highlights the strong impact of the underlying language model architecture and pretraining data on performance, even among transformer-based models.

For CamemBERT-bio, learnable fusion methods outperformed the simpler baselines that combine the CLS and entity embeddings through simple fusion operations. The Gated Fusion approach achieved the highest F1 score (39.5\%), while the Weighted Average method obtained the best AUC (75.8\%). These results suggest that incorporating entity-level information in a structured and adaptive manner enhances predictive performance, and that dynamically weighting entity contributions per patient yields the most effective representation in this case. This tendency is not applicable when using DrBERT embeddings, as the best scores in F1 and AUC are obtained when averaging the embeddings. Overall, the results consistently show that incorporating entity-level representations leads to improved performance compared to using the CLS embedding alone.
\vspace{-1em} 
\subsection{Structured data methods}

\begin{table}[!ht]
\centering
\begin{tabular}{l|cccc}
\hline
\textbf{Methods}       & \textbf{P}     & \textbf{R}     & \textbf{F1}    & \textbf{AUC}            \\ \hline
Reference & 24.80 & 63.74 & 35.67 & 76.18          \\ \hline
Baseline  & 26.24 & 69.38 & \textbf{38.02} & \textbf{79.04} \\ \hline
\end{tabular}%
\caption{Performance of structured-data only models. The best F1 and AUC results are bolded.}
\label{tab:struc-results}
\end{table}

Table~\ref{tab:struc-results} shows the results of the two logistic regressions tested on the dataset. Both structured-data models achieved strong discriminative performance, with AUC values exceeding those obtained on text-only methods. The baseline outperforms the reference model, which relies on a method tested on mortality prediction. As both have been trained on the reduced subset of 41 variables, the results indicate that, after a strong feature selection, a simpler model can perform better than an elaborated pipeline. In general, these results confirm that structured variables alone carry useful information for the task, as highlighted in previous work.

To better understand the baseline model, we analyzed the mean logistic regression coefficients across folds. The strongest positive associations with 3-month mortality include \textbf{age} and \textbf{elevated urea levels}, consistent with established risk factors reflecting advanced age and renal dysfunction in heart failure patients. Inflammatory and cardiac injury markers such as CRP and high-sensitivity troponin also contribute positively to mortality risk. In contrast, treatment variables including renin–angiotensin system inhibitors (ARA II) and beta-blockers show negative coefficients, suggesting a protective association aligned with guideline-recommended therapies. Overall, the most influential predictors correspond to clinically plausible risk factors, supporting the face validity of the structured model.

\subsection{Multimodal methods}

\begin{table*}[!ht]
\centering
\begin{tabular}{l|cccc|cccc}
\hline
\multirow{2}{*}{\textbf{Methods}}                                 & \multicolumn{4}{c}{\textbf{CamemBERT-bio}}               & \multicolumn{4}{c}{\textbf{DrBERT}}                      \\
                                                  & \textbf{P}     & \textbf{R}     & \textbf{F1}             & \textbf{AUC}            & \textbf{P}     & \textbf{R}     & \textbf{F1}             & \textbf{AUC}            \\
\hline                                                  
Baseline (CLS + All)       & 27.84 & 63.29 & 38.65          & 78.32          & 26.62 & 39.49 & 31.67          & 69.26          \\
Baseline (CLS + Sel.) & 28.70 & 66.94 & 40.15          & {\ul 81.94}    & 27.21 & 41.10 & 32.64          & 70.11          \\
\hdashline
Gated Fusion (Sel.)      & 48.99 & 48.43 & \textbf{48.48} & 81.46          & 36.14 & 48.75 & 41.11          & 75.22          \\
Dual-XAttn                              & 35.14 & 31.48 & 33.16          & 75.75          & 37.47 & 28.18 & 32.10          & 75.59          \\
Gated Dual-XAttn                        & 44.99 & 42.29 & 42.72          & 80.28          & 46.12 & 48.78 & \textbf{46.84} & {\ul 79.65}    \\
Bi-Gated Dual-XAttn          & 44.15 & 43.11 & {\ul 43.18}    & 78.24          & 50.50 & 43.92 & {\ul 46.49}    & \textbf{79.68} \\
\hdashline
Late fusion (Avg.)                             & 87.14 & 8.88  & 16.05          & 79.59          & 44.15 & 10.87 & 16.09          & 68.05          \\
Late fusion (Stack)                            & 62.77 & 29.09 & 39.54          & \textbf{83.79} & 41.43 & 26.24 & 31.01          & 75.32       \\  
\hline
\end{tabular}%
\caption{Performance of multimodal models. The best F1 and AUC scores are bolded, and the second best are underlined}
\label{tab:multimodal-results}
\end{table*}

Table~\ref{tab:multimodal-results} shows the results of multimodal models combining text embeddings and structured variables. Integrating both modalities systematically improved performance compared to unimodal baselines in terms of F1 and most of AUC scores for CamemBERT-bio. For both CamemBERT-bio and DrBERT, models using learnable fusion mechanisms consistently achieved the best F1 and AUC scores, regardless of the presence of an attention component. Interestingly, the top-performing approaches, considering the F1, Gated Fusion with selected structured variables for CamemBERT-bio (F1 48.5\%) and Gated Dual Cross Attention for DrBERT (F1 46.9\%), both leverage entity-level embeddings, highlighting the crucial contribution of fine-grained clinical information to multimodal prediction performance. It is also notable that the performance gap between DrBERT and CamemBERT-bio, which was more pronounced in the text-only setting, narrows considerably in the multimodal configuration. For CamemBERT-bio, the Late Fusion (Stack) method achieved the highest AUC (83.8\%), suggesting that even simple post-hoc ensemble methods can enhance the model's discrimination.

Paired significance testing across the five cross-validation folds confirms that multimodal late fusion significantly improves discrimination compared to both the best structured-only model (Baseline) and the best text-only model (Gated fusion) (see Appendix~\ref{app:statistical-sig}). In contrast, while structured-only models achieve higher mean AUC than text-only models, this difference shows only a trend toward significance and does not reach the conventional 0.05 threshold. 

Overall, these results demonstrate that fusing structured and unstructured data yields substantial improvements, particularly when entity-level information is incorporated. Learnable or attention-based fusion strategies consistently outperform static concatenation or averaging approaches, highlighting the benefit of jointly modeling interactions across modalities and semantic entity types. Paired significance provides statistical evidence that clinical text alone is insufficient for reliable mortality prediction, and that multimodal fusion offers the most robust discrimination performance.

\subsection{LLM-based methods}

\begin{table*}[!ht]
\setlength{\tabcolsep}{3pt}
\centering
\begin{tabular}{l*{4}{|rrr}}
\hline
\multirow{3}{*}{\textbf{Modality}} & \multicolumn{6}{c}{Qwen2.5-7B-Instruct}                                & \multicolumn{3}{c}{Mistral-7B-Instruct} & \multicolumn{3}{c}{MedGemma-4B-it} \\
                          & \multicolumn{3}{c}{Greedy} & \multicolumn{3}{c}{Constrained} & \multicolumn{3}{c}{Greedy}              & \multicolumn{3}{c}{Greedy}         \\
                          &\multicolumn{1}{c}{\textbf{P}} &\multicolumn{1}{c}{\textbf{R}} &\multicolumn{1}{c|}{\textbf{F1}} &\multicolumn{1}{c}{\textbf{P}} &\multicolumn{1}{c}{\textbf{R}} &\multicolumn{1}{c|}{\textbf{F1}}  &\multicolumn{1}{c}{\textbf{P}} &\multicolumn{1}{c}{\textbf{R}} &\multicolumn{1}{c|}{\textbf{F1}} &\multicolumn{1}{c}{\textbf{P}} &\multicolumn{1}{c}{\textbf{R}} &\multicolumn{1}{c}{\textbf{F1}}   \\
\hline
Text                      & 13.23   & 58.00   & \textbf{21.54}  & 10.92        & 90.73        & 19.50       & 15.85       & 67.21       & \textbf{25.65}       & 41.38      & 19.35     & \textbf{26.37}     \\
Structured                & 11.93   & 59.09   & 19.85  & 10.99        & 100.00          & 19.80       & 0.00           & 0.00           & 0.00           & 10.94        & 99.19       & 19.70       \\
Multimodal                & 11.10   & 98.59   & 19.95  & 11.00        & 100.00          & \textbf{19.82}       & 12.75       & 86.64       & 22.23       & 12.03        & 84.68       & 21.06      \\
\hline
\end{tabular}%
\caption{Performance of LLMs for binary mortality prediction across data modalities and decoding strategies. ``Text'' uses the raw clinical notes, ``Structured'' uses tabular patient data, and ``Multimodal'' combines both sources in a multimodal prompt.}
\label{tab:LLMs-res}
\end{table*}

Table~\ref{tab:LLMs-res} reports the performance of LLMs across text, structured and multimodal prompts. Overall, it shows that LLMs struggle with mortality prediction compared to supervised models, with consistently low F1 scores across modalities. Text-based prompts lead to better predictions than structured-only inputs, suggesting that LLMs extract more useful prognostic information from clinical reports than from raw tabular values. Multimodal prompts yield the highest recalls, along with very low precision. This effect comes from systematic single-class generation rather than genuine predictive ability. For instance, Qwen with constrained decoding predicts only the positive class, with high token probabilities (often >0.70), whereas Mistral on structured data outputs only the negative class, yielding a recall of 0 and an F1 of 0. This underlines the difficulty of enforcing binary decisions with LLMs, especially when the output does not clearly expose how the model arrives at class 0 or class 1, making it unclear whether the model is reasoning or just defaulting.
\paragraph{Decoding strategy impact:} The comparison between greedy and constrained decoding reveals differences for Qwen, where recall varies across modalities. Under constrained decoding, the model predominantly outputs class 1 with high associated probabilities, yet it remains unclear whether these outputs genuinely reflect the clinical instruction or merely result from random token selection within a restricted search space. In contrast, greedy decoding makes it easier to verify instruction adherence, since extra generated tokens immediately reveal whether the model has followed the requested output format, for instance, outputs such as \textit{Class:1} or \textit{the patient belongs to class 0}.
\paragraph{Instruction adherence:} The degree of instruction adherence varies across models and modalities. Qwen often fails to consistently follow the requested output format: in addition to the predicted class, it frequently continues generating supplementary content such as variable names, explanations, or textual analysis, an issue particularly pronounced in the structured and multimodal settings. Mistral exhibits the opposite behavior depending on the modality. For raw text, the model strictly complies with the instruction and returns a single class label, whereas with structured inputs, it tends to produce lists of variable names and values, making its final decision difficult to interpret. In the multimodal configuration, the model most often generates a clean class output. MedGemma has a perfect adherence to the constraint, consistently producing only the class label across all modalities. These discrepancies highlight that instruction-following ability is not limited to the model type, but is also strongly dependent on the modality of the prompt and its context.

Overall, across all LLMs, the highest F1-scores are consistently obtained when using text-only prompts, whereas multimodal prompts yield intermediate performance, and structured-only inputs generally result in the lowest scores. This pattern suggests that 
LLMs are more effective at exploiting information expressed in clinical text than at reasoning over structured variables, although the overall predictive performance remains low across modalities.

\subsubsection{LLM Prediction Using Textified Structured Data}
To investigate whether the limited performance of structured-only prompts is due to the format of the data or to its intrinsic predictive content, we converted the tabular variables into textual summaries using a predefined template.

\begin{table}[!ht]
\centering
\begin{tabular}{l|ccc}
\hline
\textbf{Model} & \textbf{P} & \textbf{R} & \textbf{F1} \\
\hline
Qwen 7B it     & 10.78      & 90.48      & 19.27       \\
Mistral 7B it  & 10.92      & 87.85      & 19.42       \\
MedGemma 4B it & 10.81      & 96.77      & \textbf{19.45}      \\
\hline
\end{tabular}%
\caption{Performance of LLMs when structured variables are transformed into textual descriptions using a templated representation.}
\label{tab:struc-text-res}
\end{table}

Table~\ref{tab:struc-text-res} reports the results of this experiment. For Qwen and MedGemma, F1-scores slightly decreased compared to the structured-only prompts; however, the generated outputs, particularly for Qwen, were clearer and more interpretable than when using raw structured data. 
This suggests that LLMs show difficulty in leveraging structured clinical information for this task, whether presented as text or raw tabular variables, to extract prognostic signals.
The effect is most pronounced for Mistral, which previously generated exclusively the negative class. Using textified structured data, Mistral now produces coherent class predictions, leading to a marked improvement in F1 from 0 to 19.42. This highlights that, for some models, transforming structured variables into text can improve adherence to instructions and output interpretability, even though the overall predictive gain remains modest.

\section{Conclusion}

In this work, we studied short-term mortality prediction in heart failure patients using French clinical notes and structured data. We show that integrating entity-level embeddings with CLS representations improves text-based models, and that multimodal fusion, especially with learnable or attention-based mechanisms, achieves the best overall performance. In contrast, LLMs perform poorly; they are best with text-only prompts and exhibit inconsistent behavior with structured or multimodal inputs, limiting their reliability for clinical prediction. Overall, supervised multimodal transformer models enriched with entity-level information offer the most effective approach for this task.
\section{Limitations}

This study has several limitations.
First, the dataset originates from a single French hospital and a specific clinical setting, which may limit the generalizability of our findings to other institutions, healthcare systems, or patient populations. Mortality rates, documentation practices, and structured variable distributions may vary across hospitals and countries. External validation on multi-center cohorts and different linguistic contexts is therefore necessary before clinical deployment. \\
Second, the quality of the entity-level representations depends directly on the NER annotations and the underlying schema. Extraction errors, incomplete coverage, or schema design choices may introduce bias into the learned representations and influence downstream performance. \\
Third, our LLM experiments are restricted to relatively small models (4B–7B parameters) due to hardware constraints typical of hospital environments. While this reflects realistic deployment conditions, it may limit reasoning capabilities compared to larger-scale LLMs. \\
Fourth, supervised masked language Transformer models are constrained by shorter token limits than generative LLMs, which can result in truncation of long clinical notes and potential loss of contextual information, although more recent model variants partially mitigate this issue. \\
Fifth, the evaluation of LLM outputs remains challenging. Current text-based metrics and decoding strategies do not reliably capture true clinical reasoning or interpretability. Small variations in decoding may lead to disparate predictions, and the absence of standardized, reliability-oriented evaluation metrics~\cite{qiu2025quantifying} complicates the assessment of clinical relevance. \\
Finally, despite improved predictive performance, the multimodal models provide limited interpretability, which remains a critical requirement for clinical adoption.

\section{Ethical Consideration}

This study is based on retrospective electronic health record (EHR) data collected within a secure hospital research environment. All data were de-identified prior to analysis, and no directly identifying patient information was accessible during model development. Data access was restricted to authorized personnel and conducted in accordance with institutional and national regulations governing health data use.

Mortality prediction models raise important ethical considerations. Such systems should not be interpreted as deterministic tools, but rather as decision-support mechanisms that assist clinical judgment. Any clinical deployment would require prospective validation, clinician oversight, and clear communication of model uncertainty.

Finally, the large language models evaluated in this work were tested strictly in a research context. Generative models can produce confident but incorrect predictions, and their outputs may vary depending on prompt formulation. They are not suitable for autonomous clinical use without extensive external validation and regulatory review.
\section{Acknowledgements}
This work was financially supported by ANR PREDHIC (ANR-21-CE23-0039). 

\section{Bibliographical References}\label{sec:reference}
\bibliographystyle{lrec2026-natbib}
\bibliography{lrec2026-example}

\section{Language Resource References}
\label{lr:ref}
\bibliographystylelanguageresource{lrec2026-natbib}
\bibliographylanguageresource{languageresource}
\nocitelanguageresource{*}

\section{Appendices}
\subsection{Structured Data Features}
\label{app:features}

The structured-only and multimodal models use the 41 structured variables extracted from the electronic health record, provided in Table~\ref{tab:structured_variables}.

\begin{table*}[t]
\centering
\small
\begin{tabularx}{\textwidth}{>{\bfseries}l X}
\toprule
Feature Type & Feature Names \\
\midrule

Demographics &
ADMIN\_AGE, ADMIN\_T4 \\

Biological Measurements &
BIO\_UREE, BIO\_BNP, BIO\_hco3a, BIO\_CRP, BIO\_GGT, BIO\_pha, BIO\_TROPOHS, BIO\_BILI, BIO\_CCMH, BIO\_TSH, BIO\_LDL, BIO\_POT, BIO\_ASAT, BIO\_PLAQ, BIO\_CK, BIO\_PAL, BIO\_VGM, BIO\_sao2a, BIO\_RBC, BIO\_kul \\

Vital Signs / Clinical Indicators &
INF\_pas, INF\_delta, SIGNES\_Si dyspnée \\

Diagnoses / Comorbidities &
DIAG\_Cerebrovascular Disease, DIAG\_Weight Loss, DIAG\_Maladie\_hepatique, DIAG\_Obesity, DIAG\_Renal Disease \\

Medications &
MED\_DIURETIQUES – DIURETIQUE NON EPARGNEUR POTASSIUM – DIURETIQUE THIAZIDIQUE; 
MED\_RENINE-ANGIOTENSINE – ARA II – AUTRE; 
MED\_RENINE-ANGIOTENSINE – IEC; 
MED\_BETABLOQUANTS; 
MED\_ANTIDIABETIQUES, INSULINES EXCLUES – SULFAMIDES; 
MED\_ANTIBACTERIENS A USAGE SYSTEMIQUE; 
MED\_ANTITHROMBOTIQUE – ANTICOAGULANT – HEPARINE \\

Procedures / ECG / Admission Information &
ECG\_Si QRS fin, ECG\_Si Pacemaker, MVT\_ENTREE\_TRANSFERT, TRANS\_TRANSFUSION\_SANG \\

\bottomrule
\end{tabularx}
\caption{Structured variables used in the structured-only and multimodal models.}
\label{tab:structured_variables}
\end{table*}

\subsection{Statistical Significance}
\label{app:statistical-sig}

Table~\ref{tab:significance_auc} reports paired statistical comparisons of ROC-AUC across cross-validation folds for all model pairs.

\begin{table*}[!ht]
\centering
\begin{tabular}{lccc}
\hline
\textbf{Comparison} & $\boldsymbol{\Delta}$ (Mean AUC) & \textbf{95\% CI} & \textbf{p-value} \\
\hline
Late Fusion vs  Baseline (Structured data model) & +0.0427 & [0.0203, 0.0696] & 0.0409 \\
Textual Gated fusion vs  Baseline (Structured data model) & $-0.0450$ & [$-0.0730$, $-0.0169$] & 0.0524 \\
Late Fusion vs Textual Gated fusion & +0.0904 & [0.0724, 0.1072] & 0.0009 \\
\hline
\end{tabular}
\caption{Paired significance analysis of ROC-AUC across 5 cross-validation folds. 
$\Delta$ denotes the difference between mean AUC values of the two compared models (Model A -- Model B). 
Confidence intervals are obtained via bootstrap resampling over folds; p-values correspond to paired t-tests.}
\label{tab:significance_auc}
\end{table*}

\end{document}